\documentclass{article}
\usepackage{spconf,amsmath,graphicx}
\usepackage{stfloats,framed}
\usepackage{hyperref}

\usepackage{caption}
\usepackage{subcaption}
\usepackage[belowskip=-.0pt,aboveskip=-.0pt]{caption}

\title {{ A Unified Uncertainty-aware exploration: Combining Epistemic and Aleatory  Uncertainty  }}
\name{ Parvin Malekzadeh $^{\star \dagger}$ \thanks{$^{\star}$ Corresponding author}  \thanks{ This Project was partially supported by the Innovation for Defence Excellence and Security (IDEaS) program, Canada.}\qquad   Ming Hou $^{\ddagger}$ \qquad Konstantinos N. Plataniotis$^{\dagger}$ \qquad \qquad \qquad \qquad \qquad \qquad \qquad \qquad \qquad}
\address{{\small  $^{\dagger}$ The Edward S. Rogers Sr. Department of Electrical and Computer Engineering, University of Toronto, ON, Canada  \qquad\qquad\qquad\qquad \qquad \qquad\qquad \qquad \qquad \qquad \qquad} \\
			   {\small $^{\ddagger}$ Defence Research and Development Canada (DRDC) Toronto Research Centre, ON, Canada \qquad \qquad\qquad \qquad \qquad\qquad\qquad \qquad \qquad \qquad\qquad\qquad }}
\usepackage{microtype}
\usepackage{amsmath}
\usepackage{multirow}
\usepackage{amsthm}
\usepackage{amssymb }
\usepackage{multirow}
\usepackage{amsmath,tabularx}
\usepackage{bm}
\usepackage{float}
\usepackage{amsfonts}
\usepackage{flushend}
\usepackage{geometry}
\usepackage{mathtools}
\usepackage[dvipsnames]{xcolor}
\usepackage{algorithm, algpseudocode}
\def\mS{\mathcal{S}}
\def\mA{\mathcal{A}}
\def\SR{\text{UUaE}}

\def\mS{\mathcal{S}}

\def\mA{\mathcal{A}}

\def\k{_{k}}

\def\li{^{(l,i)}}

\def\bt{\bm{\theta}}
\def\h{\bm{h}}

\def\v{\bm{v}}

\def\m{\bm{m}}

\def\s{\bm{s}}
\def\pk{_{k-1}}

\def\l{^{(l)}}

\def\t{_t}
\def\nt{_{t+1}}
\def\v{\bm{v}}

\newcommand\myeq{\stackrel{\mathclap{\normalfont\mbox{D}}}{=}}
\makeatletter
\newcommand{\multiline}[1]{%
  \begin{tabularx}{\dimexpr\linewidth-\ALG@thistlm}[t]{@{}X@{}}
    #1
  \end{tabularx}
}
\begin{document}
\ninept
\maketitle
\begin{figure*}[b!]
\begin{framed}
Accepted paper \href{https://doi.org/10.1109/ICASSP49357.2023.10095594}{https://doi.org/10.1109/ICASSP49357.2023.10095594}. ©2023 IEEE.
\end{framed}
\end{figure*}
%%%
\begin{abstract}
Exploration is a significant challenge in practical reinforcement learning (RL), and uncertainty-aware exploration that incorporates the quantification of epistemic and aleatory uncertainty has been recognized as an effective exploration strategy. However, capturing the combined effect of aleatory and epistemic uncertainty for decision-making is difficult. Existing works estimate aleatory and epistemic uncertainty separately and consider the composite uncertainty as an additive combination of the two.  Nevertheless,  the additive formulation leads to excessive risk-taking behavior, causing instability. In this paper, we propose an algorithm that clarifies the theoretical connection between aleatory and epistemic uncertainty, unifies aleatory and epistemic uncertainty estimation, and quantifies the combined effect of both uncertainties for a risk-sensitive exploration. Our method builds on a novel extension of distributional RL that estimates a parameterized return distribution whose parameters are random variables encoding epistemic uncertainty. Experimental results on tasks with exploration and risk challenges show that our method outperforms alternative approaches.
% a novel algorithm that unifies estimation of b,oth aleatory and epistemic uncertainty and c.  Our algorithm builds on
%network,
\end{abstract}
\begin{keywords}
Belief, Exploration, Uncertainty
\end{keywords}
\vspace{-.1in} \section{Introduction}
\label{sec:intro}\vspace{-.1in}
Reinforcement learning (RL) has been widely applied in various domains, such as robotics and autonomous vehicles, by creating decision-making agents that interact with their environments and receive reward signals. Despite the successes in many domains, poor sample efficiency during learning makes deploying RL agents in real-world applications unfeasible~\cite{yang2021exploration,lockwood2022review}. This challenge becomes even more severe when sample collection is expensive or risky. One promising approach for improving sample efficiency is uncertainty-aware exploration, which uses uncertainty in both the agent and the environment~\cite{lockwood2022review,chen2022efficient}. The uncertainty in the agent, known as epistemic uncertainty, arises from the agent's imperfect knowledge about the environment. As the agent learns more about its environment, the epistemic uncertainty decreases. The uncertainty in the environment, referred to as aleatory uncertainty, originates from intrinsic stochasticity and persists even after the environment's model has been learned~\cite{moerland2017efficient}. While incorporating and quantifying both uncertainties in supervised learning has been explored~\cite{campbell2022robust,fang2022integrating}, this problem in RL is not yet well-understood~\cite{yang2021exploration,lockwood2022review}. Considering uncertainty for decision-making leads to a risk-sensitive policy, where being pessimistic or optimistic about aleatory uncertainty creates a risk-averse or risk-seeking policy, respectively. A risk-seeking policy makes an agent consistently revisit states with high risk due to the irreducibility of aleatory uncertainty, degrading performance~\cite{yang2021exploration}. Many algorithms perform optimistic decision-making based solely on epistemic uncertainty, but these approaches can lead to excessive risk-seeking behavior since they cannot identify areas with high aleatory uncertainty. Thus, beyond estimating the epistemic uncertainty, capturing the aleatory uncertainty also prevents the agent from exploring areas with high randomness.
\vspace{.02in}  \\
\textbf{Related Work:} Recently, several methods have been developed for estimating epistemic or aleatory uncertainty separately in RL. Distributional RL~\cite{dabney2018distributional,dabney2020distributional} is a popular approach that learns the distribution of returns and is used to measure aleatory uncertainty. On the other hand, methods such as bootstrap sampling~\cite{lee2021sunrise, lee2022offline, bai2022pessimistic}, Monte Carlo dropout~\cite{hiraoka2022dropout, wu2021uncertainty}, and Bayesian posterior~\cite{turchetta2020robust,chen2022efficient, malekzadeh2020mm, salimibeni2020distributed} are commonly used to estimate epistemic uncertainty.
However, simultaneously estimating both types of uncertainty and incorporating them for sample-efficient learning is challenging. Some algorithms, such as those in~\cite{moerland2017efficient, tang2018exploration, keramati2020being}, incorporate epistemic uncertainty into aleatory uncertainty estimation by sampling parameters that define the return distribution. The agents then explore actions with high aleatory uncertainty. However, due to the non-reducibility of aleatory uncertainty during training, forcing the agent to choose actions with high aleatory uncertainty can lead to excessive risk-taking behavior and instability~\cite{clavier2022robust,mavor2022stay}. This is because despite reducible epistemic uncertainty, aleatory uncertainty cannot be decreased during training.
Mavrin et al.~\cite{mavrin2019distributional} proposed a method to suppress the effect of aleatory uncertainty by applying a decay schedule, but this method does not consider epistemic uncertainty. Some other works~\cite{eriksson2022sentinel,clements2019estimating,mai2022sample} estimate epistemic and aleatory uncertainty separately and combine them using a weighted sum. However, due to the reducibility of epistemic uncertainty and non-reducibility of aleatory uncertainty during training, this additive formulation underestimates the integrated effect of epistemic uncertainty.
Other approaches, such as those in~\cite{chen2022efficient,nikolov2018informationdirected}, use information-directed sampling to avoid the negative impact of aleatory uncertainty by balancing instantaneous regret and information gain. However, these information-directed approaches are slow and computationally intractable for practical problems with large state or action spaces since they require learning the transition dynamics of the environment.
\vspace{.02in} \\
\textbf{Contributions:}
To capture and quantify the integrated effect of aleatory and epistemic uncertainty, we formalize an analytical connection between them and propose a unified uncertainty estimation algorithm. Specifically: \\
1. By maintaining a belief distribution over a set of possible parameters defining a return distribution, we propose a so-called belief-based distributional RL framework that reveals a formal relationship between aleatory and epistemic uncertainty. The proposed belief-based distributional RL scheme also unifies the estimation of both uncertainties and provides a basis to extend existing distributional RL methods that currently only quantify epistemic uncertainty to learn both types of uncertainty. \\
2. By modelling the belief by a mixture of Dirac deltas, we derive novel learnable features based on Moment-Generating functions (MGFs) corresponding to the belief. Approximating the high-dimensional belief with a mixture of Dirac deltas benefits us with efficient computation and easier handling of the non-linear propagation of the belief in the derived belief-based distributional RL scheme. The MGF features provide rich statistics of the belief and allow us to leverage well-explored distributional RL algorithms in our non-trivial belief-based distributional RL system. \\
3. Finally, we present a unified exploration method that considers the estimated composite uncertainty for exploration. Our proposed exploration strategy paves the way for future research on designing exploratory policies considering the combined effect of aleatory and epistemic uncertainty. We apply this exploration strategy to challenging tasks such as Atari games and an autonomous vehicle driving simulator and demonstrate that our method achieves substantial improvements in stability and sample efficiency compared to existing frameworks that only consider aleatory uncertainty, epistemic uncertainty, or an additive combination of both uncertainties.
%OOOOOOOOOOOOOOOOOOOOOOOOOOOOOOOOOOOOOOOOOOOOOOOOOOOOOOOOO
  \vspace{-.1in}\section{Preliminaries} \label{sec:background}
%OOOOOOOOOOOOOOOOOOOOOOOOOOOOOOOOOOOOOOOOOOOOOOOOOOOOOOOOO
%In what follows, we represent scalar variables by Non-bold letter (e.g., $X$), vectors by lowercase bold letter (e.g., $\x$), matrices by capital bold letter (e.g., $\X$), and transpose of matrix $\X$ by $\X^T$.
%
%=========================================================
  \vspace{-.1in}
\textbf{Markov Decision Process (MDP):}
%=========================================================
We model the agent-environment interaction with a stationary MDP specified by $\mathcal{M}=(\mS, \mA, P, \gamma, r)$, where $\mS$ and $\mA$ are state and action spaces, respectively. $P(\s\nt|\s\t,a\t)$ is the unknown transition dynamics and provides the probability of successor state $\s\nt$ given the present state $\s\t$ and action $a\t$ at time step $t$. $\gamma\in (0,1)$ is the discount factor, and $r(\s\t,a\t)$ is the unknown scalar reward function given $(\s\t,a\t)$. Commonly, within the RL context, the objective of the agent is to find a policy $\pi: \mS \rightarrow \mA$ that maximizes the expected return $ Q^{\pi}(\s\t, a\t)=\mathbb{E} [\sum_{k=t}^{\infty}\gamma^{k-t} r(\s\k,a\k)]$, where $\s\k \sim P(.| \s\pk, a\pk)$ and $a\k =\pi(\s\k)$.
\vspace{.03in}  \\
\textbf{Distributional RL:}
The objective in distributional RL~\cite{dabney2018distributional} is to learn the probability distribution of return $Z^{\pi}(\s\t,a\t)=\sum_{k=t}^{\infty}\gamma^{k-t} r(\s\k,a\k)$ by solving the distributional Bellman equation:
\vspace{-.1in}
{\small
\begin{equation}
Z^{\pi}(\s\t, a\t) \myeq  r(\s\t,a\t)+ \gamma Z^{\pi}(\s\nt, a\nt), \label{Eq:Bellman}
%&& \!\!\!\!\!\! \s' \sim \text{Pr}(.|\s,a), a' \sim \pi(.|\s),
\end{equation}} \normalsize \par  \vspace{-.1in}
\noindent{where} $\myeq$ represents distributional equality.  By approximation of the return distribution through a neural network with parameters $\bm{\psi}$ and given samples $(\s\t, a\t, r\t=r(\s\t,a\t), s\nt)$, the problem of return distribution estimation can be formulated as a minimization problem as:
\vspace{-.05in}{
\begin{equation}
\!\!\!\! \bm{\theta}\!= \!\arg\min_{\bm{\theta}'} J \! \left(Z^{\pi}_{\bm{\theta}'}(\s\t, a\t),  r\t \!+\! \gamma Z^{\pi}_{\bm{\theta}^-}(\s\nt, a\nt)\right), \label{Eq:Distr_Bellman}
%&& \!\!\!\!\!\! \s' \sim \text{Pr}(.|\s,a), a' \sim \pi(.|\s),
\end{equation}} \par  \vspace{-.05in}
\noindent{where} $J$ is a statistical distance, $\bm{\theta}^-$ is the latest value of $\bt$, and $a\nt =\pi(\s\nt)$. \vspace{.01in} \\ 
 The distribution of $Z^{\pi}_{\bt}(\s_t,a_t)$, $p(Z^{\pi}_{\bt}(\s_t,a_t))$, captures aleatory uncertainty and is induced by the stochasticity in the transition dynamics $P$. However, epistemic uncertainty accounts for the uncertainty about parameter $\bt$ and is provoked by limited data availability for learning $\bt$.
%==========================================
  \vspace{-.1in}\section{PROBLEM FORMULATION} \label{sec:formulation} \vspace{-.1in}
 Distributional RL approaches typically involve approximating the return distribution by parameterizing it as a categorical distribution over finite atoms~\cite{chen2022efficient,nikolov2018informationdirected, bellemare2017distributional}. However, these algorithms often require domain-specific knowledge, such as the number of atoms and the bounds of the support, to reduce approximation error. Choi et al.~\cite{choi2019distributional} proposed a solution to these issues by modeling the return distribution with a mixture of Gaussians, also known as a Gaussian mixture model (GMM). GMMs can approximate any distribution to arbitrary accuracy by adjusting the number of mixtures and naturally accommodate the return distribution's multi-modality, resulting in more stable learning. In our case, the distribution of $Z^{\pi}$ can be represented using $L$ Gaussian mixtures with parameters ${\bt}=\{\bt\l \}_{l=1}^L$ as:
\vspace{-.1in} {\small
\begin{eqnarray}
\lefteqn{\!\!  Z^{\pi}_{{\bt}}(\s\t,a\t)} \label{Eq:GMM} \\
&& \!\!\!\! \! \sim \sum_{l=1}^{L} w_{\bt\l}(\s\t,a\t)~\mathcal {N}(u_{\bt\l}(\s\t,a\t),\sigma^2_{\bt\l}(\s\t,a\t)), \nonumber
% \vspace{-.1in}
\end{eqnarray}}\normalsize \par  \vspace{-.11in}
\noindent{where} $w_{\bt\l}(\s\t,a\t)$, $u_{\bt\l}(\s\t,a\t)$, and $\sigma^2_{\bt\l}(\s\t,a\t)$ are the $l^{\text{th}}$ mixture weight, mean, and variance function, respectively. Unlike existing distributional RL algorithms where the parameters of the return distribution are assumed to be deterministic values, in our case, due to the unknown MDP model, the agent cannot determine the distribution parameters with complete reliability. Thus, there exists epistemic uncertainty about the parameters $\bt$. We can think of $Z^{\pi}$ as a random variable $Z^{\pi}_{\bm{\Theta}}$ whose distribution parameter is random variable ${\bm{\Theta}\!=\!\{ \bm{\Theta}\l\}_{l=1}^L}$  taking values on ${ {\bt}\!=\!\{ \bm{\theta}\l\}_{l=1}^L \! \in \! \mathcal{R}^{L \times D} }$.
% To address the challenge of estimating the distribution of a random variable of random variables, a common approach, as in~\cite{moerland2017efficient,clements2019estimating,mai2022sample}, is to estimate the empirical return distribution by randomly sampling different values of $\bt$. However, this approach often leads to poor out-of-sample performance~\cite{derman2020distributional}.
%OOOOOOOOOOOOOOOOOOOOOOOOOOOOOOOOOOOOOOOOOOOOOOOOOOOOOOOOO
  \vspace{-.1in} \section{PROPOSED METHOD}\label{sec:AKF-SR}
%OOOOOOOOOOOOOOOOOOOOOOOOOOOOOOOOOOOOOOOOOOOOOOOOOOOOOOOOO
  \vspace{-.1in} This section describes our proposed approach, called Unified Uncertainty-aware Exploration ($\SR$), which is presented in Algorithm~\ref{algo:summary}. $\SR$ comprises three main components: (1) a belief-based distributional RL scheme that integrates both aleatory and epistemic uncertainty, (2) an estimation method that learns the composite uncertainty, and (3) an exploration strategy that exploits the composite uncertainty for exploration. \vspace{-.13in}
  \subsection{Belief-based distributional RL} \vspace{-.05in}
%%%%%%%%%%%%%%%%%%%%%%%
\begin{algorithm}[t]
\caption{\textproc{The proposed framework.}}
\label{algo:summary}
\begin{algorithmic}[1]
%\State \textbf{Learning Phase:}
\State {  \multiline{ \textbf{Input:}  $L$: size of GMM, $K$: number of delta mixtures, $\gamma$: discount factor, $M$: size of MGF features.}}
%\State {  \multiline{ \textbf{Initialize:} \\ $\{ \bt_0^a \}_{a \in \mA}$: parameter vector of the reward function \\ $\{ \F_{0}^a\}_{a \in \mA}$: parameter matrix of the dynamics \\ $\{\S_0^a\}_{a \in \mA}$: posterior covariance of $\{ \F_{0}^a\}_{a \in \mA}$ \\ $\{ \bm{\Pi}_{0}^a\}_{a \in \mA}$: posterior covariance of $\{ \bt_0^a \}_{a \in \mA}$ \\ $\{\bm{\phi}_0(\s)\}_{\s \in \mS}$: feature vector } }
%\For{$n=1,2,..., N_{\text{episode}}$}
\State {  \multiline{ \textbf{Initialize:} $\s_0$: initial state, $\bm{\phi}^-$: parameters of previous belief, $\bm{\phi}$: parameters of current belief.}}
\For {$t=0,1, ...$}
\State \!\!\!\! Compute MGF features $\tilde{\m}(\bm{\phi})$.
% \State \!\!\!\! \multiline{ Choose $a\t\!=\! \arg\max_{a'}(\mathbb{E}[Z_{\tilde{\m}(\bm{\phi})}(\s\t,a')]\!-\! Var[Z_{\tilde{\m}(\bm{\phi})}(\s\t,a')])$.}
\State \!\!\!\! \multiline{Take action $a\t\!=\! \arg\max_{a'}\{\mathbb{E}[Z_{\tilde{\m}(\bm{\phi})}(\s\t,a')]\!-\! Var[Z_{\tilde{\m}(\bm{\phi})}(\s\t,a')]\}$, observe $\s\nt$ and $ r\t$.}
\State \!\!\!\! \multiline{Choose  $a\nt\!\!=\! \arg\max_{a}\{\mathbb{E}[Z_{\tilde{\m}(\bm{\phi^-})}(\s\nt,a)]\!-\!\! Var[Z_{\tilde{\m}(\bm{\phi^-})}(\s\nt,a)]\}$.}
\State \!\!\!\! \multiline{Compute { ${L(\bm{\phi})}\!\!=\!\! JTD(Z_{\tilde{\m}(\bm{\phi})}(\s\t, a\t),r\t+\gamma Z_{\tilde{\m}(\phi^-)}(\s\nt,a\nt))$.}\normalsize}
\State \!\!\!\! \multiline{ Update $\bm{\phi}$ as {$\bm{\phi} \leftarrow \bm{\phi}- \nabla L(\bm{\phi})$.}\normalsize}
\State \!\!\!\! \multiline{Update $\bm{\phi}^-$ as $\bm{\phi}^- \leftarrow \bm{\phi}.$\normalsize}
%\State { $\bm{\phi}\t(\s) \leftarrow \bm{\phi}\pt(\s)$ .}
\!\!\!\! \EndFor
\end{algorithmic}
\end{algorithm} 
%%%%%%%%%%%%%%%%%%%%%%%%%%%%%%%%%%%%
To avoid the ambiguity of ``the distribution of a random variable of a random variable $Z^{\pi}_{\bm{\Theta}}$", we represent epistemic uncertainty in the form of a belief distribution $b$ over a set of plausible parameters $\bt$, i.e., $b(\bt)=p(\bm{\Theta}= \bt )$. Assuming that $\bm{\theta}\l$ for $l=\{1,2,...,L \}$  are distributed independently, we have $b({\bt})=\prod_{l=1}^L b\l(\bt\l)$, where $b\l(\bt\l)=p(\bm{\Theta}\l=\bt\l)$. As the agent does not know the exact value of $\bt$, it must be able to learn the return distribution from the belief $b$. By defining ${Z_{b}^{\pi}} = \mathbb{E}_{b(\bt)}[Z_{\bt}^{\pi}]$ and ${Z_{b^-}^{\pi}} = \mathbb{E}_{b(\bt^-)}[Z_{\bt^-}^{\pi}]$ and by taking the expectation of Eq.~\eqref{Eq:Bellman} with respect to $b(\bt)$  and  $b(\bt^-)$, the so-called belief-based distributional Bellman equation is derived as
\vspace{-.1in} {\
\begin{eqnarray}
Z_{b}^{\pi}(\s\t, a\t) \myeq  r(\s\t,a\t)+ \gamma Z_{b^-}^{\pi}(\s\nt, a\nt), \label{Eq:Distr_Bellman_belief}
%&& \!\!\!\!\!\! \s' \sim \text{Pr}(.|\s,a), a' \sim \pi(.|\s),
\end{eqnarray}}\par  \vspace{-.09in}
\noindent{where}
 \vspace{-.07in}{\
\begin{eqnarray}
\lefteqn{\!\!\!Z^{\pi}_{{b}}(\s\t,a\t)} \label{Eq:GMM2} \\
&& \!\!\!\! \!\!\!  \sim \sum_{l=1}^{L} w_{b\l}(\s\t,a\t)~\mathcal {N}(u_{b\l}(\s\t,a\t),\sigma^2_{b\l}(\s\t,a\t)).  \nonumber
\end{eqnarray}\par  \vspace{-.1in}
%%%
} \normalsize
\noindent{As parameters of $p(Z^{\pi}_{{b}})$ reflect the belief over $\bt$, we refer to $p(Z^{\pi}_{{b}})$ as the belief-based return distribution. The derived belief-based distributional Bellman equation} provides a basis for incorporating epistemic uncertainty into the aleatory uncertainty learning processes of current distributional RL methods.
\vspace{-.12in}
\subsection{ Unified uncertainty estimation} \vspace{-.05in}
 Parameterizing $p(Z^{\pi}_{{b}})$ directly with the belief $b= \{b\l \}_{l=1}^L$  is non-trivial. To feed $b$ as the parameters of $p(Z^{\pi}_{{b}})$ to a network, we require to
extract sufficient features of $b$.  To that end, we first need to characterize $b$. We begin by modelling $b\l$ as a mixture of $K$ Dirac deltas with parameters ${\bm{\phi}\l = \{(\h\li, \alpha\li )\}_{i=1}^K \in \mathcal{R}^{K \times 2}}$ describing locations and weights of Dirac deltas. \\
To pivot the non-trivial belief-based return distribution estimation problem into a previously-solved return distribution estimation problem, which can be solved through Eq.~\eqref{Eq:Distr_Bellman}, we introduce a feature extraction method for $b\l$ based on MGF of $\bm{\Theta}\l$. MGF of a random variable is a computationally efficient alternative specification of its probability distribution~\cite{johnson2019statistics}. The MGF of $\bm{\Theta}\l$ with belief $b\l$ for a fixed vector $\v$ is given by $MGF_{\bm{\Theta}\l}(\v)=\sum_{i=1}^K \alpha \li e^{\v^T \h\li}$. We represent $b\l$  with the feature vector ${\m}(\bm{\phi}\l)\!  = \{ \mathbb{E}[(\bm{\Theta}\l)^j] \}_{j=1}^M$, where ${\mathbb{E}[(\bm{\Theta}\l)^j] \!  =\!  \mathbb{E}[\bm{\Theta}\l \otimes ...  \otimes \bm{\Theta}\l] \! \in \!\mathcal{R}^{D^j}}$ is the $j$-th order moment ($j$-way tensor) of $\bm{\Theta}\l$ achieved from the $j$-th order
derivative of the MGF at ${\v\! =\!\bm{0}}$. Thus, the total features for ${b \! =\! \{b\l \}_{l=1}^L}$  are given in ${\tilde{\m}(\bm{\phi})\! =\!\{ {{\m}(\bm{\phi}^{l}) \}_{l=1}^L} \! \in \! \mathcal{R}^{L \times \sum_{j=1}^M \! D^j} }$. Using Dirac deltas for modelling $b$ has several benefits: (1) A mixture of Dirac deltas can be considered as a GMM with zero (co)variances. Hence, like GMMs, a mixture of deltas can approximate any probability distribution by adjusting the number of mixtures~\cite{choi2019distributional}. (2) The MGF corresponding to a mixture of deltas exists on an open interval around $\v=\bm{0}$, and higher-order moments of $\bm{\Theta}$ can be computed efficiently. (3) The moments of $\bm{\Theta}$ are differentiable with respect to $\bm{\phi}$; thus, $\bm{\phi}$ can be directly optimized via gradient descent during back-propagation.  \\
Lastly, by substituting $\tilde{\m}(\bm{\phi})$ for $b$ in Eq.~\eqref{Eq:Distr_Bellman_belief}, we obtain:
%%%
\vspace{-.1in}{
\begin{eqnarray}
Z^{\pi}_{\tilde{\m}(\bm{\phi})}(\s\t, a\t) \myeq  r(\s\t,a\t)\! + \!\gamma Z^{\pi}_{\tilde{\m}(\bm{\phi}^-)}(\s\nt, a\nt). \label{Eq:Distr_Bellman_belief2}
%&& \!\!\!\!\!\! \s' \sim \text{Pr}(.|\s,a), a' \sim \pi(.|\s).
\end{eqnarray}}  \par \vspace{-.05in}
% \noindent{where}
% \vspace{-.2in} {
% \begin{eqnarray}
% \lefteqn{\!\!\!\!\!\!\!\!\!\!\!\!\! Z^{\pi}_{\tilde{\m}(\bm{\phi})}(\s\t,a\t) \sim \sum_{l=1}^{L} w_{\m(\bm{\phi}\l)}(\s\t,a\t)} \label{Eq:GMM3} \\
% && \,\,\,\,\,\,\,\,\, ~\mathcal {N}(u_{\m(\bm{\phi}\l)}(\s\t,a\t),\sigma^2_{\m(\bm{\phi}\l)})(\s\t,a\t). \nonumber
% \end{eqnarray}} \par \vspace{-.1in}
% %%%
%
%%%%
\vspace{-.016in}  To learn the parameter vector $\bm{\phi}$, we utilize a mixture density network~\cite{choi2019distributional} with a total parameter vector $\tilde{\m}(\bm{\phi})$.  Our network takes in $[\s\t, a\t]^T$ as input and produces $3 \times L$ outputs $\{ w_{\tilde{\m}(\bm{\phi}\l)}\}_{l=1}^L$, $\{u_{\tilde{\m}(\bm{\phi}\l)}\}_{l=1}^L$, and  $\{\sigma_{\tilde{\m}(\bm{\phi}\l)}\}_{l=1}^L$. The network must minimize the statistical distance between two GMMs: $Z^{\pi}_{\tilde{\m}(\bm{\phi})}(\s\t, a\t)$ and $r\t+\gamma Z^{\pi}_{\tilde{\m}(\bm{\phi}^-)}(\s\nt,a\nt)$.
The Wasserstein distance and Kullback-Leibler divergence, which are commonly used in distributional RL literature~\cite{chen2022efficient,dabney2020distributional,nikolov2018informationdirected}, between two GMMs are analytically intractable. Thus, we utilize the recently proposed Jensen-Tsallis Distance (JTD)~\cite{choi2019distributional}, which has a closed form between two GMMs and provides unbiased sample gradients.
The gradient of the JTD loss function with respect to $\bm{\phi}$ is calculated as a matrix multiplication between the gradient of the loss with respect to the network parameters $\tilde{\m}(\bm{\phi})$ and Jacobian matrix of the MGF-feature extractor function $\tilde{\bm{m}}$ as  ${\nabla L(\bm{\phi})}=  {\mathbb{J}_{ \tilde{\m}}^T(\bm{\phi})} {\nabla L(\tilde{\m}(\bm{\phi}))}.$ %%%
%%%%
\vspace{-.12in} \subsection{Composite uncertainty-aware exploration} \vspace{-.05in}  
% The composite mean and variance of  $Z^{\pi}_{\tilde{\m}(\bm{\phi})}$ \normalsize  are calculated as ${\mathbb{E}[Z^{\pi}_{\tilde{\m}(\bm{\phi})}] \!\!=\!\! \sum_{l=1}^{L} \! w_{\tilde{\m}(\bm{\phi}\l)}u_{\tilde{\m}(\bm{\phi}\l)} \normalsize \text{\, and \,} \small Var(Z_{\tilde{\m}(\bm{\phi})}) \! =\! \sum_{l=1}^{L} \! w_{{\tilde{\m}}(\bm{\phi}\l)}\sigma^2_{\tilde{\m}(\bm{\phi}\l)} }  + +\sum_{l=1}^{L} w_{\tilde{\m}(\bm{\phi}\l)} \! \left( \! u_{{\tilde{\m}}(\bm{\phi}\l)} \! - \!\! \sum_{n=1}^{L} \! w_{\tilde{\m}(\bm{\phi}\n)}u_{\tilde{\m}(\bm{\phi}\n)} \! \right) ^2 \! .$\normalsize \, 
Following common practices in RL~\cite{mavrin2019distributional,mai2022sample}, we interpret the uncertainty of a random variable as its variance (or trace of its covariance). Thus, $\tilde{\m}(\bm{\phi})$ \normalsize captures the epistemic uncertainty since $\mathbb{E}[\bm{\Theta}^2]\!=\! \text{vec}({Cov}(\bm{\Theta})) \!+\! \mathbb{E}[\bm{\Theta}]\otimes\mathbb{E}[\bm{\Theta}]$\normalsize, and the composite variance \small $Var(Z_{\tilde{\m}(\bm{\phi})})$ \normalsize is the irreducible aleatory uncertainty with the parameter  $\tilde{\m}(\bm{\phi})$ \normalsize including epistemic uncertainty.  Therefore, while attempting to maximize expected rewards $\mathbb{E}[Z_{\tilde{\m}(\bm{\phi})}]$, the agent should avoid regions with high aleatory uncertainty to avoid excessive risk-seeking behavior, i.e.,
% Following common practices in RL~\cite{mavrin2019distributional,mai2022sample}, we interpret the uncertainty of $Z^{\pi}_{\tilde{\m}(\bm{\phi})}$ as its variance. Thus, the variance of $Z^{\pi}_{\tilde{\m}(\bm{\phi})}$ measures the composite uncertainty as it integrates the epistemic uncertainty into the aleatory uncertainty via the feature vectors $\bm{\phi}$. {As the learning converges}, the epistemic uncertainty reduces and composite uncertainty decreases to the aleatory uncertainty. Thus, to prevent high-risk states, the agent should penalize actions with high aleatory uncertainty, i.e.,
%%%
\vspace{-.05in}{
\begin{equation}
\!\!a\t\!\!=\! \arg\max_{a'}\{\mathbb{E}[Z_{\tilde{\m}(\bm{\phi})}(\s\t,a')]\!-\! Var(Z_{\tilde{\m}(\bm{\phi})}(\s\t,a'))\}. \nonumber
\end{equation}}  \par
\begin{figure}[t]
\vspace{5mm}
\centering
\includegraphics[width=7.41cm,height=2.71cm]{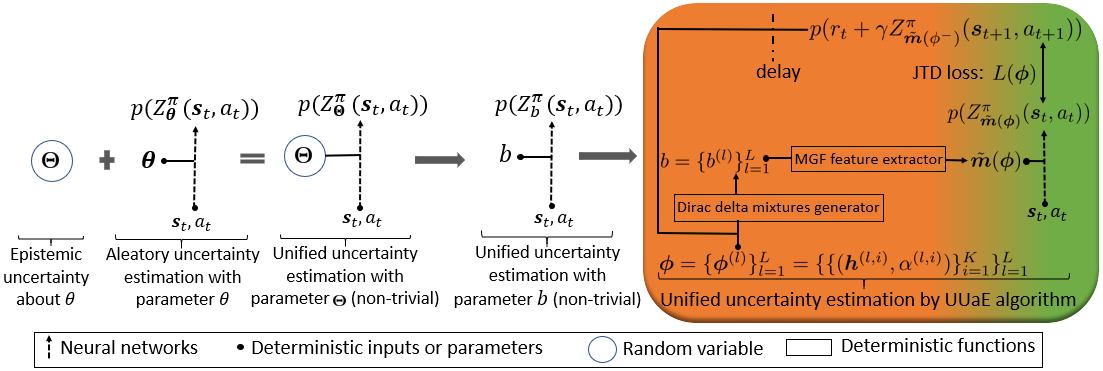} 
 \vspace{-.05in} \caption{{  Steps taken to derive our proposed $\SR$ method. Orange and green shaded areas show learning of epistemic and aleatory uncertainty in $\SR$. }} \label{Fig:1} \vspace{-.0in}
\end{figure}

 \vspace{-.07in}
%%%
Fig.~\ref{Fig:1} provides an illustration of uncertainty estimation in our proposed $\SR$ method.
%%%%%%%%%%%%%%%%%%%%%%%%%
\vspace{-.1in}\section{ EXPERIMENTS} \label{sec:results}
\begin{figure*}[!b]  \vspace{-.25in}
\begin{minipage}[b]{.32\linewidth}
  \centering
  \centerline{\includegraphics[width=4.1cm, height=3.1cm]{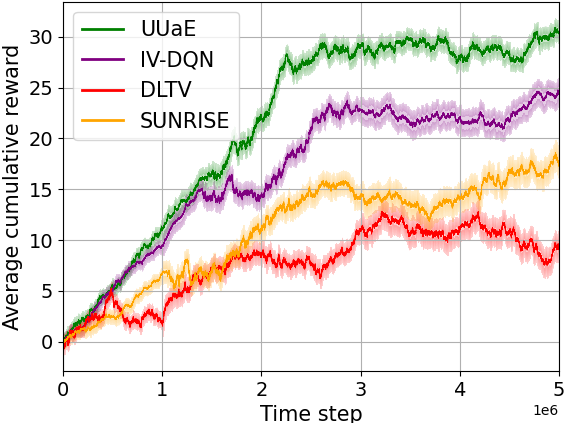}}
%  \vspace{1.5cm}
  \vspace{-.05in} \centerline{(a) Atari-Asterix}\medskip
\end{minipage}
\hfill
\begin{minipage}[b]{0.32\linewidth}
  \centering
  \centerline{\includegraphics[width=4.1cm,height=3.1cm]{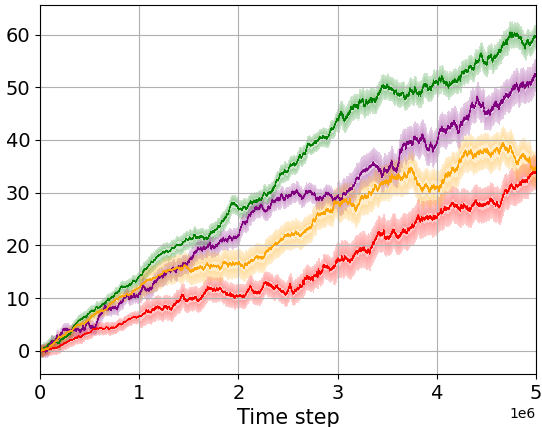}}
%  \vspace{1.5cm}
  \vspace{-.05in} \centerline{(b) Atari-Seaquest}\medskip
\end{minipage}
\hfill
\begin{minipage}[b]{0.32\linewidth}
  \centering
  \centerline{\includegraphics[width=4.1cm,height=3.1cm]{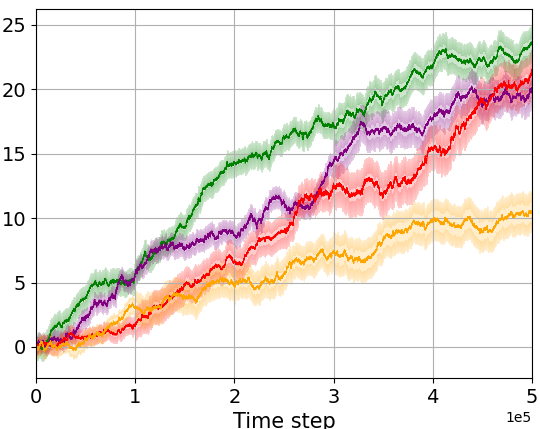}}
%  \vspace{1.5cm}
  \vspace{-.05in} \centerline{(c) Autonomous vehicle driving}\medskip
\end{minipage}
\caption{Learning curves on two Atari games and autonomous vehicle driving task across $10$ runs.}
\label{Fig:res}
\end{figure*}

\vspace{-.1in} In this section, we evaluate the performance of $\SR$ on two Atari games with sparse reward functions. To test the proposed algorithm in a more realistic setting, we also run our algorithm on an autonomous vehicle driving simulator~\cite{leurent2018environment} in a highway domain, where rewards are designed to penalize unsafe driving behavior. The sparsity and risk-sensitivity of rewards in these tasks make uncertainty-aware exploration challenging. \vspace{.02in} \\
\textbf{Baselines:} We compare the performance of $\SR$ to three recently proposed algorithms: SUNRISE~\cite{lee2021sunrise}, DLTV~\cite{mavrin2019distributional}, and IV-DQN~\cite{mai2022sample}, which respectively act based on the epistemic uncertainty, aleatory uncertainty, and additive formulation of epistemic and aleatory uncertainty.\vspace{.02in} \\
\textbf{Setup:}  We implement the baselines using their original implementations. The hyperparameters $L$, $K$, and $M$ of $\SR$ are empirically set to $L=5$, $K=10$, and $M=10$. The experiments indicated a monotonic increase in performance with higher values of $L$, $K$, and $M$. However, larger values of these hyperparameters lead to an increase in computation time. Thus, a trade-off between the computation time and performance must be made to choose these values. \vspace{.02in} \\
%which are referred to as Ra-UNAE and Rn-UNAE, respectively.  to $L=5$, $K=10$, and $M=10$
\textbf{Results:} Following common approaches in RL literature~\cite{lee2021sunrise,mavrin2019distributional,mai2022sample}, we measure the performance of algorithms in terms of the cumulative reward.  Fig.~\ref{Fig:res} presents the results averaged over $10$ random seeds, with shaded areas indicating the standard deviation. Our algorithm outperforms all three baselines across all three environments, demonstrating the significance of our approach in integrating both aleatory and epistemic uncertainties for exploration and in the effectiveness of MGF features as a summary of the belief. Additionally, our method exhibits low variance and good risk-sensitive performance, making it more stable than the baselines. In the autonomous driving task, SUNRISE's performance deteriorates substantially as it does not consider aleatory uncertainty, whereas driving safely on the highway necessitates risk-sensitive driving.
%%%%%%%%%%%%%%%%%%%%%%%%%%%%%%%%%%%%%%%%%%
%%%%%%%%%%%%%%%%%%%%%%%%%%%%%%%%%%%%%%%%%%%%
%%%%%
%%%%%
\vspace{-.1in} \section{Conclusion} \vspace{-.1in} \label{sec:conclusion}
Relying on a novel extension of the distributional RL to the so-called belief-based distributional RL, we provide a theoretical contribution for joint epistemic and aleatory uncertainty estimation. The proposed method scales up current distributional RL algorithms, which only consider aleatory uncertainty, to measure both sources of uncertainty. To efficiently explore environments with composite uncertainty, we approximate the belief as a mixture of Dirac deltas and extract features using MGFs. Our method outperforms recent alternatives and exhibits stable exploration in highly sparse and uncertain environments.
%%%%%%%%%%%%%%%%%%%%%%%%%%%%%%%%%%%%%%%%%%%%
\vfill\pagebreak
\clearpage
% -------------------------------------------------------------------------
\bibliographystyle{IEEEbib}
{\footnotesize \bibliography{strings,refs}}

\end{document}